\documentclass[12pt]{article}

\usepackage{graphicx}
\usepackage{amsmath}
\usepackage{amssymb}
\usepackage{booktabs}
\usepackage{url}
\usepackage{cite}
\usepackage[margin=1in]{geometry}

\newcommand{\hlsfml}{\texttt{hls4ml}}
\newcommand{\snntorch}{\texttt{snnTorch}}
\newcommand{\pytorch}{\texttt{PyTorch}}
\newcommand{\snnreadout}{\texttt{SNNReadout}}
\newcommand{\neurobench}{\texttt{NeuroBench}}

\begin{document}

\title{Spiking Neural Network inference \\ on FPGAs with \hlsfml}

\author{Barry M. Dillon\\
{\small{ ISRC, Ulster University, Derry, BT48 7JL, Northern Ireland}}\\
\texttt{ {\small b.dillon@ulster.ac.uk}}}

\date{}

\maketitle

\begin{abstract}
Spiking Neural Networks (SNNs) provide a naturally temporal machine-learning framework.
Their neurons maintain an internal state and propagate information through discrete spikes, enabling low-latency temporal inference.
Although SNNs are often associated with asynchronous neuromorphic processors, many scientific real-time inference systems rely on conventional synchronous field-programmable gate arrays (FPGAs) and high-level synthesis (HLS) workflows.  
In this paper we present an extension of \hlsfml{} that enables clock-driven deployment of SNNs trained in \pytorch{} onto FPGA firmware.  
We demonstrate the workflow using a dense quantised SNN trained on the Heidelberg Spiking Digits dataset where it achieves inference latencies of approximately $34\mu\text{s}$. 
We validate the generated design through software reference comparisons, HLS C simulation, HLS synthesis, export, and Vivado synthesis reports.
This work opens up the \hlsfml{} toolkit to neuromorphic computing, allowing streamlined optimisation, synthesis, and deployment of SNN models for real-time inference.
\end{abstract}


\section{Introduction}

Spiking neural networks (SNNs) \cite{ANDERSON1972197} are often described as a third generation of neural network models because they replace continuous stateless activations with event-like spikes that have an internal dynamical state that persists and changes in time \cite{Maass:1997spiking}.  
With the development of deep-learning using SNNs \cite{TAVANAEI201947,oconnor2016deepspikingnetworks} this representation is attractive for processing temporal data in hardware systems that can exploit sparse data for low power and low latency inference.  
Dedicated neuromorphic processors such as TrueNorth \cite{Merolla:2014million}, Loihi \cite{Davies:2018loihi}, and SpiNNaker \cite{Furber:2014spinnaker} illustrate the potential of large-scale event-based or spike-oriented computing systems.  
At the same time, many scientific instruments already use synchronous FPGAs for deterministic, low-latency, real-time processing.  For such systems, a clock-driven SNN implementation can enable trained spiking models to be deployed in standard firmware workflows while taking advantage of the stateful neuron and temporal readout semantics.

The \hlsfml{} project was developed to translate trained machine-learning models into HLS firmware for low-latency FPGA and ASIC implementations \cite{fastml_hls4ml,Duarte:2018ite,Schulte:2025mai}.  
It has been used extensively in scientific real-time inference applications, especially where resource and latency constraints require careful consideration in the model design, numerical precision, and hardware implementation.  
Extensions to \hlsfml{} have demonstrated how new neural-network layer classes can be added to the toolchain and characterized through end-to-end accuracy, latency, resource, and quantisation studies.  
For example convolutional layer compatibility \cite{Aarrestad:2021zos,Ghielmetti:2022ndm}, distributed arithmetic for efficient constant matrix vector multiplication (CMVM) \cite{Sun:2025}, and binary/ternary compatibility \cite{Loncar:2020hqp}.

Low-latency SNN algorithms are paving the way for new scientific applications \cite{mehonic2024roadmapneuromorphiccomputingemerging}, with works exploring their use in temporal scientific data classification \cite{10.1145/3183584.3183612}, radiation detection \cite{10.3389/fphy.2024.1334298}, vertex reconstruction in neutrino-nucleus interactions \cite{8683736}, as well as many tasks at the Large Hadron Collider (LHC) such as data filtering \cite{kulkarni2023onsensordatafilteringusing}, particle tracking \cite{Coradin_2025}, readout for hadron calorimeters \cite{lupi2025neuromorphicreadouthadroncalorimeters}, and real-time anomaly detection \cite{Dillon:2025yqe}.

This paper introduces support for SNNs in \hlsfml{} with Vivado/Vitis backends.  
Spiking neurons defined in \snntorch{} \cite{eshraghian2021training} are mapped to backend templates in \hlsfml{} and compiled to High-Level Synthesis (HLS).
The linear layers use the same \pytorch{} interface to \hlsfml{} that exists already.
The spiking neurons are stateful and contain information that persists over time until it is read out at the end of a fixed window in time, when a decision rule is applied to the SNN output.
This readout process is handled by a configurable \snnreadout{} layer that encodes this behaviour.
The current SNN implementation is clock-driven and synchronous, therefore unlike dedicated neuromorphic hardware, the algorithms do not consume lower power with sparse data inputs.  
This work allows SNNs to be optimised for real-time FPGA deployment through the exact same workflow as regular Artificial Neural Networks (ANNs).
It builds upon the already widely used \hlsfml{} framework, utilising its well-tested optimisation procedures and pipelines, and allows for implementations of hybrid spiking / non-spiking algorithms.

We test the SNN functionality in \hlsfml{} using the Spiking Heidelberg Digits (SHD) dataset \cite{Cramer_2022}, a well-known benchmark for SNNs consisting of spike-encoded audio recordings of spoken digits.
We compare two different methods for SNN readout, direct membrane potential output and spike outputs.
To compress the networks we apply a custom Quantisation Aware Training (QAT) algorithm on both the linear layers and the spiking neurons.
We then compare the HLS conversion for a range of different bit width precisions, and run a full Vitis/Vivado synthesis to obtain predictions for latency and resource usage on a xczu7ev-ffvc1156-2-e part.
Lastly we look at the \neurobench{} \cite{yik2025neurobenchframeworkbenchmarkingneuromorphic} metrics to compare the spiking activity in the networks both for different readout methods and for different bit widths.
The paper is organised as follows.
Section \ref{sec:bkg-rel-work} covers background on SNNs, FPGA deployment, and related work. 
Section \ref{sec:snn-support} details the support for SNNs that we have implemented in \hlsfml{}.
The SHD dataset and the preprocessing we use is described in Section \ref{sec:shd}. 
In Section \ref{sec:shd-analysis} we present our analysis of the SHD dataset using SNNs with a full \hlsfml{} synthesis.
Finally we conclude in Section \ref{sec:conclusions}.

\section{Background}
\label{sec:bkg-rel-work}

\subsection{Spiking neural networks \& surrogate-gradients}

SNNs compute with time-dependent neuron states and binary spike events.
In a conventional artificial neuron neuron \cite{rosenblatt1958perceptron} information is processed as
\begin{equation}
\label{eq:nnact}
y=f(wx+b),
\end{equation}
where $x$ is the input to the neuron, $y$ is the neuron output, $f$ is the nonlinear activation function, and $(w,b)$ are the learnable weight and bias terms.
So the artificial neuron is stateless, there are no internal parameters that change with time, and the nonlinear activation function is differentiable which allows for training via backpropagation.
A spiking neuron however processes information differently, for example the Leaky Integrate and Fire (LIF) neuron is described by
\begin{align}
    y_t &= \begin{cases}
0 & \text{if } u_t < u_{\text{thresh}} \\
1 & \text{if } u_t \geq u_{\text{thresh}}
\end{cases} \\
    u_{t+1} &= \beta u_t + w x_{t+1} - \beta y_t u_{\text{thresh}} + b.
\end{align}
At a timestep $t$ the neuron receives an input current $x_t$ which updates its membrane potential $u_t$, i.e. the internal state of the spiking neuron.
The membrane potential accumulates current contributions over the timesteps, with $0\leq\beta\leq1$ defining a leakage of the potential over time.
The neuron generates a spike at step $t$ if the membrane potential $u_t$ exceeds the neurons threshold potential $u_{\text{thresh}}$, at which point the membrane potential resets.
This nonlinear behaviour in the spiking is what gives SNNs their expressivity.
However unlike the artificial neurons the process is not differentiable and so directly training SNNs with gradient methods is not possible.
Surrogate-gradient methods address this by using a differentiable proxy for the spike derivative during backpropagation while retaining the spiking forward dynamics \cite{oconnor2016deepspikingnetworks,Zenke_2018,8891809}.  
The \snntorch{} package builds on this approach by exposing spiking neurons as \pytorch{} \cite{paszke2019pytorchimperativestylehighperformance} modules and providing practical tools for training SNNs with modern deep-learning workflows \cite{Eshraghian:2023lessons}.  
In this paper, \snntorch{} is used as the training frontend and the source framework for the example model.

\subsection{SNN inference}

SNNs are constructed from conventional linear layers with layers of spiking neurons separating them, much like conventional Artificial Neural Networks (ANNs) can be viewed as a sequences of linear and activation layers.
After data is passed through a layer of LIF neurons (i.e. a forward pass) their internal state changes, so that when the next item of data arrives the computation done by the layer is different.
How the internal state of the LIF neurons change is determined by the weights and biases in the linear layers, and by the spiking neuron parameters $\beta$ and $u_{\text{thresh}}$.
Both the weights and biases in the linear layer and the leakage parameter in the spiking neurons can be optimised during training via gradient descent.
SNNs are designed to stream time-series data, as soon as an input at timestep $t$ has been processed by the first layer, an input at timestep $t+1$ is ready to be processed by the first layer, and so on until the end of the fixed-window.
So spikes generated by an input at time $t$ may not have reached the readout layer by the time the SNN has begun to process the next inputs.
This is what makes SNNs naturally temporal and naturally suited for processing time-series data.
In regular ANNs the temporal dimension of the network must be constructed using recurrent connection or persistent state vectors.

For time-series classification problems, as we will consider in this paper, classification is usually done at the end of a fixed window in time.
We consider two potential readout mechanisms from the SNN: (i) neuron membrane readout from the final layer, (ii) spike-count readout from the final layer.
For neuron membrane readout the threshold $u_{\text{thresh}}$ on the final layer must be fixed large enough such that the neurons can accumulate sufficient potential without spiking, while for the spike-count readout the threshold is optimised during training.
From the neuron membranes and spike counts we can then construct scores for each time-series sample and use these for classification.
An overall accuracy score can be obtained from a simple \texttt{argmax} on the readouts.
It is also possible to obtain early classification before the fixed window has completed.
This can be done using first-to-threshold methods where the first neuron on the final layer of the SNN to reach a threshold potential or spike count determines the classification output.

\subsection{SNN accelerators \& this work}

Many works have already demonstrated that implementing SNNs on FPGA hardware can deliver low-latency inference with performances comparable with deep-learning using artifical neurons, for example on image classification with benchmark datasets such as MNIST \cite{6701396,Han2020,ZHANG2020106,9256533,9556469,nevarez2021,carpegna2022spikerfpgaoptimizedhardwareacceleration,Chen_2022,Chen_2022_2,li2023fireflyv2advancinghardware}.
The S2N2 architecture \cite{10.1145/3431920.3439283} delivers an FPGA implementation of SNNs focused specifically on streaming time-series data, and online learning for SNNs on FPGAs has also been explored in \cite{fastefficientonlinesnn}.
Recent works have pushed performance further in terms of throughput \cite{Li_2023,li2025fireflythighthroughputsparsityexploitation}, model size \cite{Aung_2023}, and latency and power \cite{10191153,10666827}.
With the popularity of deep-learning development frameworks such as \pytorch{} there has been recent interest in interfacing SNN FPGA accelerators with Python, for example via \snntorch{} making use of surrogate gradients \cite{fan2025robustopensourceframeworkspiking}, or using the Spike-Timing-Dependent Plasticity learning mechanism \cite{gautam2025neurocorexopensourcefpgabasedspiking}.
As well as accelerators for specific architectures more general frameworks for mapping SNNs to FPGAs have been developed recently, with the \texttt{spiker+} \cite{Carpegna_2025} and \texttt{QUANTISENC} \cite{matinizadeh2024fullyconfigurableopensourcesoftwaredefineddigital} tools mapping configurable multi-layer SNNs to FPGAs and ASICs through a Python interface.
However each of the tools mentioned here is a standalone framework developed specifically for SNNs.
What we are presenting here extends a general HLS-based model compiler in \hlsfml{} with SNN semantics, rather than proposing a custom framework.
This builds upon the well-developed and well-maintained tools provided through \hlsfml{}, and integrates SNN and ANN architectures into a single framework.  

\section{SNN support in \hlsfml{}}
\label{sec:snn-support}

The SNN implementation in \hlsfml{} extends the existing \pytorch{} frontend, intermediate layer representation, and Vivado/Vitis HLS template system, while retaining the usual \hlsfml{} workflow of model conversion, C simulation, HLS synthesis, and export \cite{Duarte:2018ite,Fahim:2021hls4mlCodesign,Schulte:2025hls4mlPlatform}.  
Unlike dedicated neuromorphic hardware we are not developing an asynchronous event-driven SNN algorithms, the goal is to implement a clock-driven SNN algorithm that fits the current \hlsfml{} workflow incorporating stateful, timestep-by-timestep SNN inference.

\subsection{Clock-driven execution}
\label{subsec:clock-driven-execution}

Clock-driven means that the neuron potential updates, dense layer computations, and readout updates are scheduled by ordinary HLS pipelines and streams.  
In the current interface convention, one call to the generated top function processes one timestep of one input sequence.  
For an input window of length $T$, inference for one independent example therefore consists of $T$ sequential top-function invocations.  
The SNN layers contain a static internal state in the generated C++ implementation that persists across top-function calls until the configured readout window boundary is reached.

This execution model differs from native asynchronous neuromorphic hardware.  
In an event-routed implementation, computation and communication may be triggered by spike events and may avoid work when no events are present.  
In the present implementation, by contrast, the HLS design advances once per input timestep according to the FPGA clock and interface handshakes.  
Nevertheless, the model semantics remain spiking and stateful where hidden IF/LIF neuron layers update membrane variables over time, emit binary spikes, and reset state only at the configured sequence boundary.  
This makes the implementation appropriate for fixed-window SNN inference in FPGA workflows.

\subsection{Frontend support}
\label{subsec:pytorch-snntorch-frontend}

The frontend support is implemented through the \pytorch{} conversion path.  
The converter treats \snntorch{} modules as leaf modules during tracing, which allows users to write ordinary \pytorch{} models containing \texttt{snntorch.Leaky} modules.  
Initial support is only for the \texttt{Leaky} neuron, which is the dominant spiking neuron used in SNN architectures, but future work will extend beyond this.  
During conversion, a \texttt{Leaky} module is mapped either to a \texttt{LIFNeuron} layer or, when its leak parameter $\beta$ is equal to one, to an \texttt{IFNeuron} (integrate and fire) layer.  

For \texttt{Leaky} neurons, the converter reads the current values of the \snntorch{} parameters at conversion time.  
The supported reset mechanisms for the neuron potentials after spiking are \texttt{subtract} and \texttt{zero}.  
Both the threshold parameter $u_{\text{thresh}}$ and the leak parameter $\beta$ may be either scalar or a per-neuron vector of length \texttt{n\_out}.
If these parameters are set to learnable in \snntorch{} they automatically take the per-neuron vector form.
Scalar values are represented as compile-time configuration constants in the generated design, while per-neuron values are represented as parameter arrays.  
This distinction is useful because many SNNs use a shared threshold or decay value, but trainable or heterogeneous neuron parameters should still be representable without changing the layer interface.

\subsection{Intermediate SNN layer representation}
\label{subsec:snn-ir-layers}

The converted model graph contains three new SNN-specific layer types: \texttt{IFNeuron}, \texttt{LIFNeuron}, and \texttt{SNNReadout}.  
The \texttt{IFNeuron} layer stores the number of input/output channels, the fixed-window size, threshold information, reset mode, and membrane-state precision.  
Its generated HLS function implements the update
\begin{equation}
    u_{t+1} = u_t + x_{t+1}, \qquad
    y_{t+1} = \Theta(u_{t+1} - u_{\text{thresh}}),
\end{equation}
followed by either subtractive reset, $u_{t+1} \leftarrow u_{t+1}-u_{\text{thresh}}$, or zero reset, $u_{t+1}\leftarrow 0$, if a spike occurs.  
The \texttt{LIFNeuron} layer extends this with a decay parameter,
\begin{equation}
    u_{t+1} = \beta u_t + x_{t+1}, \qquad
    y_{t+1} = \Theta(u_{t+1} - u_{\text{thresh}}),
\end{equation}
with the same reset choices.  
In both cases this is for an individual neuron where each may have its own learned $\beta$.
The output tensor has the same shape as the input tensor except for the final dimension, which is set by \texttt{n\_out}.
In the common dense-SNN use case this final dimension is unchanged by the neuron layer.

The generated C++ templates provide both array-based and \texttt{hls::stream}-based versions of the IF and LIF kernels.  
In each kernel, membrane state is stored in a static array of length \texttt{n\_out}, this array is completely partitioned by HLS pragmas so that neuron updates can be unrolled across the output dimension when resources allow.  
A static timestep counter tracks progress through the fixed window.  
When the counter reaches \texttt{window\_size}, the counter and membrane array are reset, allowing the next independent sequence to start from zero state.

\subsection{Readout modes and decision rules}
\label{subsec:snn-readout}

The \snnreadout{} layer is an \hlsfml{} readout layer with a corresponding \pytorch{} layer module.  
In \pytorch{} this layer is simply an identity operation, so it does not change training-time computation.  
During conversion, however, the layer is mapped to the \texttt{SNNReadout} layer in the \hlsfml{} graph.  
This layer defines how timestep-wise class signals are accumulated into a sequence-level decision, so it is key in the orchestration of the SNN inference.
\snnreadout{} supports two output modes, \texttt{spike} and \texttt{membrane}.  
In \texttt{spike} mode, the readout counts the discrete spikes output by the final LIF layer over the whole course of the readout window.
The supported spike-count decision rules include spike-count argmax, first-to-threshold, threshold-then-argmax, and binary-logit.  
The spike-count argmax rule emits the class with the largest accumulated count.  
The threshold-based rules use a configurable class-count threshold.  
For binary classifiers, the binary-logit rule emits the signed count difference between class~1 and class~0.
In \texttt{membrane} mode, the readout is intended to be placed directly after the final dense layer, rather than after a final spiking neuron.  
The readout owns a separate membrane state for each class and updates it as
\begin{equation}
    m_{t+1} = \beta_{\mathrm{ro}} m_t + z_{t+1},
\end{equation}
where $z_{t+1}$ is the dense current supplied to class $i$ at timestep $t+1$, and $\beta_{\mathrm{ro}}$ is a scalar readout decay.  
No threshold or reset-on-spike is applied within the readout membrane update.  
The supported membrane decision rules are \texttt{argmax\_membrane} and \texttt{binary\_logit} where the latter emits the membrane difference $m_1-m_0$ for binary classification.  


\subsection{Window boundaries and state reset}
\label{subsec:window-boundary-semantics}

The current generated kernels implement fixed-window reset semantics.  
During conversion, the \snnreadout{} layer provides the value of \texttt{window\_size}, which is propagated to the converted IF and LIF layers.  
Each stateful SNN layer maintains an internal timestep counter.  
At the end of the configured window, after the final timestep has contributed to the neuron update and readout decision, the layer state is cleared for the next independent sequence.  
For neuron layers this clears the neuron state while for readout layers it clears either the spike-count array or the readout neuron array.

This behaviour means that a compiled SNN model is stateful across calls to \texttt{hls\_model} \texttt{.predict()}.  
The evaluation code will therefore call the compiled model exactly once per timestep and pass exactly \texttt{window\_size} timesteps for each independent sequence.  
Stray single-timestep calls made before a sequence evaluation will advance the internal state and can change the next prediction.  
This is unlike a purely feed-forward non-spiking model, where each call is normally independent.
Native variable-length SNN inference is left as future work.
True AXI-stream packet-boundary support would require top-level interface and writer support for propagating packet sideband information such as \texttt{TLAST} into the SNN layer kernels.  

\subsection{Fixed-point precision and numerical fidelity}
\label{subsec:snn-precision}

The SNN layers use the ordinary \hlsfml{} precision system.
Users may configure output result precision and the SNN-specific threshold, LIF decay ($\beta$), and neuron potential precisions using HLS precision types.  
In the generated templates, hidden-neuron potential state and membrane-mode readout accumulation are stored using the configured membrane types.

We implement our own fake-quantised \pytorch{} model for QAT whose rounding and saturation behaviour matches the \hlsfml{} configuration.  
The SHD example uses explicit Vitis HLS fixed-point rounding and saturation modes, for example \texttt{ap\_fixed<8,4,\allowbreak AP\_RND\_CONV,\allowbreak AP\_SAT>}.  
The software-to-HLS agreement reported in section~\ref{subsec:software-to-hls} compares the compiled \hlsfml{} model to the fake-quantised, stateful \pytorch{} reference under the same timestep and readout semantics.

\subsection{Current limitations}
\label{subsec:snn-current-limitations}

The SNN support in \hlsfml{} is currently for only LIF neurons.
While this is the most prevalent and fundamental building block of SNNs and is enough to construct SNNs for many applications, it will be interesting in future work to add support for other types of spiking neurons.
For example a key addition would be the recurrent LIF neuron, \texttt{RLeaky} in \snntorch{}, which allows feedback from spike outputs of neurons in the same layer to affect the state.
Other limitations include the fixed-window nature of the readout, and the wholly clock-driven nature of the inference.
Fixed-window readout could be relaxed if we allow decisions to be propagated from the readout layer once the spike-count or output neuron-potentials reach a certain threshold.
In future we can also include gated execution for the forward pass through the network, such that only computations in which a spike is emitted are performed on the hardware.
The gating will not reduce inference latency, but for sparse inputs and sparse firing in the intermediate layers, it could reduce power consumption.
In the SHD example we use the \neurobench{} tool to track the spiking sparsity in the intermediate layers, however the \hlsfml{} pipeline with Vitis synthesis will not give us estimates of power savings due to sparsity in the spiking.

\section{SHD dataset}
\label{sec:shd}

The SHD dataset \cite{Cramer_2022} consists of recordings of digits $0\!-\!9$ spoken in both English and German, therefore we have $20$ classes.
The authors of the dataset used an audio-to-spike conversion procedure, inspired by the neurophysiology of the cochlea, to convert each audio sample into a spike train.
The neurophysiological model consists of $700$ channels to approximate the spiking activity in the inner ear.
Each event contains a label for the class, and a list of spikes and the time at which it arrives in each channel.

We start by binning the spikes for each event in time using a bin size of $10$~ms and a window size of $T=140$ which corresponds to a $1.4$~s window.  
Spikes outside the $1.4$~s window are discarded.  
The 700 input channels are then reduced to 70 input features by averaging over consecutive groups of 10 source channels in each time bin.
This is physically motivated and is also done in \cite{Cramer_2022}.

\begin{figure}
\centering
\includegraphics[width=\linewidth]{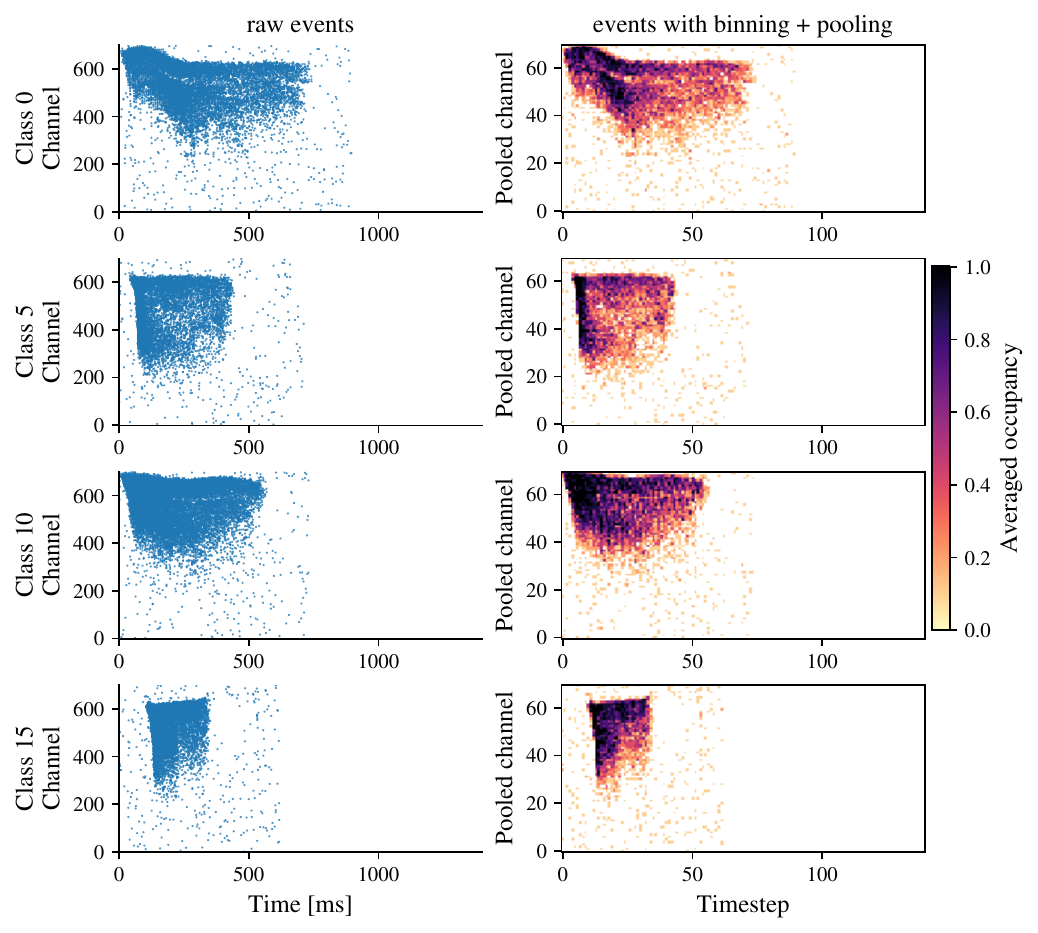}
\caption{Representative SHD test samples shown as raw 700-unit event rasters and the 140-by-70 pooled occupancy heatmaps used as model inputs.}
\label{fig:shd-data-images}
\end{figure}

\begin{figure}
\centering
\includegraphics[width=0.45\linewidth]{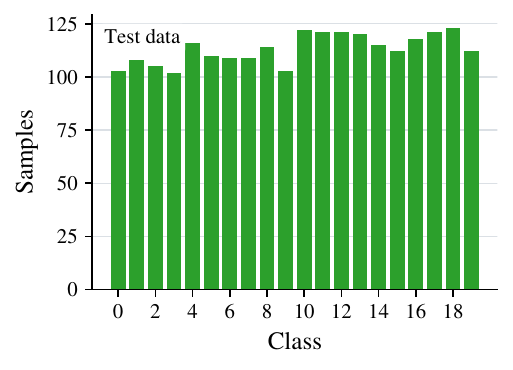}
\includegraphics[width=0.45\linewidth]{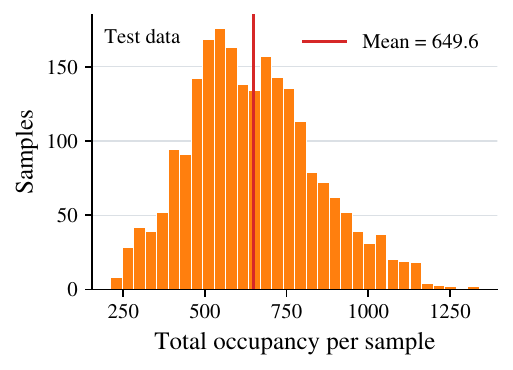}
\caption{On the left we show the held-out test-set class balance, and on the right we show the distribution of total occupancy per test sample.}
\label{fig:shd-data-summary}
\end{figure}

The preprocessing produces a dense tensor of pooled spike occupancy values rather than a strict binary channel-selection tensor.  
The SHD dataset has a predefined train / test split, with $8156$ training events and $2264$ test events.
The speakers in the training and testing events are different, with the purpose being that performance on the test set checks the generalisability of the classification algorithm across different speakers.
In this work we further split the training dataset into $7156$ training events and $1000$ validation events for early stopping, and retain the same $2264$ test set for testing.
Because the test set contains held-out speakers it is not identically distributed with the training and validation data, therefore we naturally expect differences in validation and testing accuracies and report both in the results of the analysis. 
A summary of the preprocessing is presented in Table~\ref{tab:dataset-preprocessing}, and visualisations of the dataset properties are shown in Figure~\ref{fig:shd-data-images} and Figure~\ref{fig:shd-data-summary}.

\begin{table}[htbp]
\centering
\caption{SHD preprocessing and split used for all experiments.}
\label{tab:dataset-preprocessing}
\begin{tabular}{lc}
\toprule
Quantity & Value \\
\midrule
Timesteps per sample & 140 \\
Sample duration & 1.4 s \\
Temporal bin width & 10 ms \\
Input channels & 70 pooled from 700 \\
Input representation & averaged spike occupancy \\
Training samples & 7156 \\
Validation samples & 1000 \\
Test samples & 2264 \\
Mean input occupancy per test sample & 649.6 \\
Number of classes & 20 \\
\bottomrule
\end{tabular}
\end{table}

\section{SNN classification of SHD in \hlsfml{}}
\label{sec:shd-analysis}

We consider a very simple network architecture with a $70$-dim input layer, a single hidden spiking layer with $64$ LIF spiking neurons, and a $20$ dimensional output (readout) layer.
With membrane readout we simply read off the states on the output layer of the network after the fixed-window has ended, while for spike output the final layer has $20$ LIF spiking neurons and we count the spikes accumulating from each neuron over the duration of the window.
All LIF neurons in the SNN have trainable leak ($\beta$) and threshold ($u_{\text{thresh}}$) parameters and subtract-reset dynamics.  During training, dropout with probability 0.3 is applied to the hidden spike vector before the final dense layer, and dropout is disabled for evaluation, QAT export, and HLS conversion.  
We find that dropout helps with generalisation from training and validation data to the testing data with the SHD dataset.
For the readout we test both the membrane output and the spike output representation.

At each timestep a dense layer maps the 70-dimensional pooled input to hidden neurons.  
The LIF neuron potentials are updated by these inputs, and they emit spikes when the potentials exceed the threshold $u_{\text{thresh}}$, followed by a subtractive reset.  
The hidden decay parameter is initialized to $\beta=0.75$, the thresholds are initialised to $1.0$, and both are trainable.  
The second dense layer maps emitted spikes to 20 output neurons.  
In the membrane-readout configuration, these outputs are accumulated by \snnreadout{} and the predicted class is the largest final readout value. 
In the spike-count configuration, a second spiking layer is inserted before the readout and the readout accumulates output spike counts over the duration of the fixed window.

\subsection{Training \& QAT}

The models are first trained in full 32 bit floating point (FP32) precision and then fine-tuned with QAT.  
The training is done with \snntorch{} which uses \pytorch{} on the backend, and the gradient descent methods rely on surrogate gradients for the spiking neurons.
We use a $10$-fold cross validation procedure that cycles through training ($7156$ events) and validation ($1000$ events) data, allowing us to implement early stopping on the training, and giving us uncertainty estimates on the results.
FP32 runs use Adam with learning rate $10^{-3}$, cosine annealing to $10^{-4}$, a batch size of $64$, and a cross-entropy loss.
Runs are configured for at most 500 epochs and use early stopping on validation accuracy with patience $50$ and minimum improvement $5\times 10^{-4}$.  
QAT starts from the best FP32 checkpoint, uses Adam with learning rate $3\times10^{-4}$, cosine annealing to $5\times10^{-5}$, a batch size of $64$, the same cross-entropy loss, and early stopping with patience 25.  
The QAT forward pass quantises inputs, weights, bias, neuron parameters, membrane updates, dense outputs, and readout membrane state using straight-through estimators.

\subsection{Spike vs membrane readouts}

\begin{figure}[h]
\centering
\begin{minipage}{0.5\linewidth}
\centering
\includegraphics[width=\linewidth]{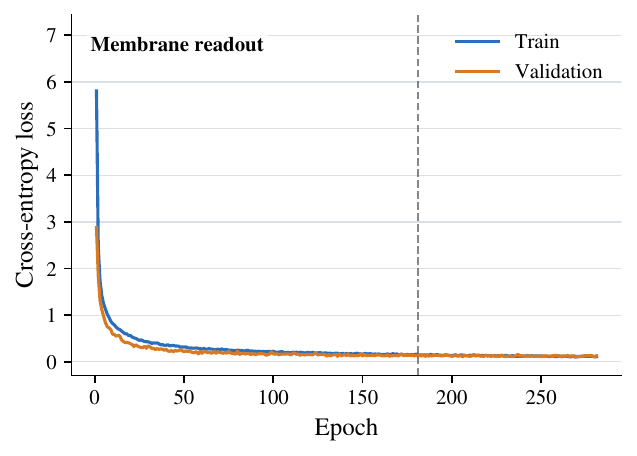}
\end{minipage}\hfill
\begin{minipage}{0.5\linewidth}
\centering
\includegraphics[width=\linewidth]{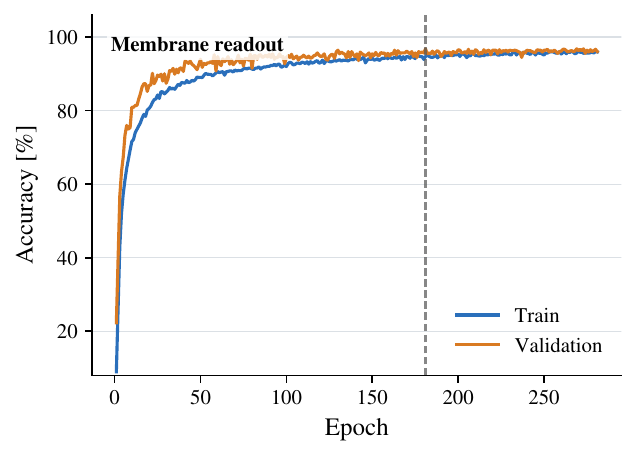}
\end{minipage}
\vspace{0.5em}
\begin{minipage}{0.5\linewidth}
\centering
\includegraphics[width=\linewidth]{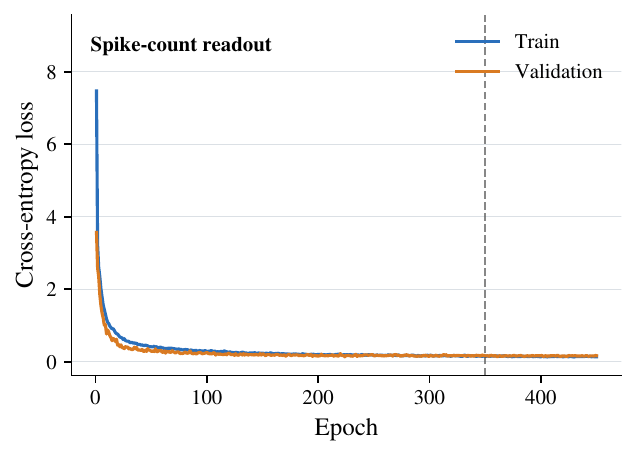}
\end{minipage}\hfill
\begin{minipage}{0.5\linewidth}
\centering
\includegraphics[width=\linewidth]{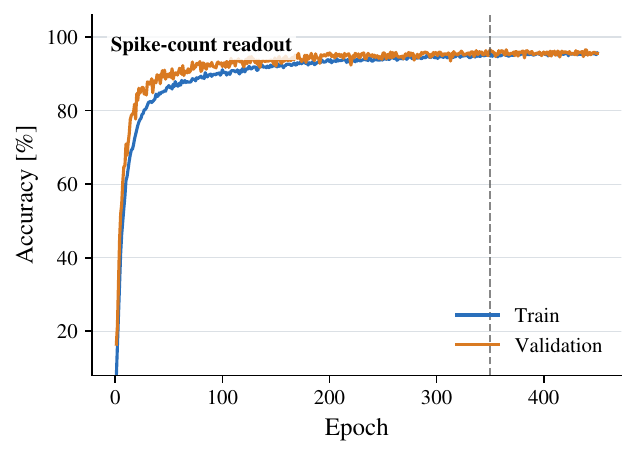}
\end{minipage}
\caption{Accuracy and loss curves for one fold of the cross-validation from both the membrane-readout and spike-readout training.}
\label{fig:memspike-training-curves}
\end{figure}

The first result we discuss is the comparison between SNNs trained with full FP32 precision using spike vs membrane outputs.
In general we find that the membrane readout SNNs result in a larger accuracy than the spike-readout SNNs.
This is likely due to sparse firing on the output layer essentially leading to a lower resolution output than we get from the membrane potential.
In Figure~\ref{fig:memspike-training-curves} we show how the accuracy and loss converge over the epochs for one fold of the cross-validation.
We see that both readout types evolve similarly, but when looking at the mean and spread of the accuracies in Table~\ref{tab:fp32-readout-comparison} we can see that the membrane readout SNNs perform better.
For this reason we will go forward with only the membrane readout models for HLS conversion in the next sections.
We also see the performance gap between the validation and test accuracies here.
As discussed previously, this is due to these datasets being drawn from different underlying distributions (i.e. different speakers), and the same gap is seen in other works.

\begin{table}[h]
\centering
\footnotesize
\begin{tabular}{lccc}
\toprule
Readout & Validation accuracy & Test accuracy & Test loss \\
\midrule
Membrane   & $96.79 \pm 0.64\%$ & $75.81 \pm 1.00\%$ & $1.195 \pm 0.175$ \\
Spike count & $94.56 \pm 1.42\%$ & $73.94 \pm 2.09\%$ & $1.372 \pm 0.224$ \\
\bottomrule
\end{tabular}
\caption{Comparison of membrane and spike-count readout SNNs with full FP32 precision. Accuracy and test loss are reported as mean and standard deviation over 10 folds.}
\label{tab:fp32-readout-comparison}
\end{table}

\subsection{Software-to-HLS fidelity}
\label{subsec:software-to-hls}

Here we want to test how well the SNN outputs are preserved when going from models trained in \pytorch{} to implementation in HLS via \hlsfml{}.
Starting from the FP32 membrane readout networks from the previous section, we continue with QAT until the models converge.
We consider several precisions for the HLS implementation:  \texttt{ap\_fixed<10,4>}, \texttt{ap\_fixed<12,4>}, \texttt{ap\_fixed<16,4>}, and \texttt{ap\_fixed<24,4>}.
After QAT we then run the networks through \hlsfml{} with the following config choices:
\begin{itemize}
    \itemsep0pt
    \item backend: \texttt{Vitis}
    \item FPGA part: \texttt{xczu7ev-ffvc1156-2-e}
    \item \texttt{IOType = io\_stream}, for timestep-by-timestep SNN inference
    \item fixed-point arithmetic: convergent rounding (\texttt{AP\_RND\_CONV}) and saturation (\texttt{AP\_SAT})
    \item strategy: \texttt{Resource}
    \item input layer reuse factor:  $\mathrm{RF}=7$
    \item hidden layer reuse factor:  $\mathrm{RF}=8$
    \item target clock period: $16$~ns.
\end{itemize}

\begin{figure}[h]
\centering
\begin{minipage}{0.48\linewidth}
\centering
\includegraphics[width=\linewidth]{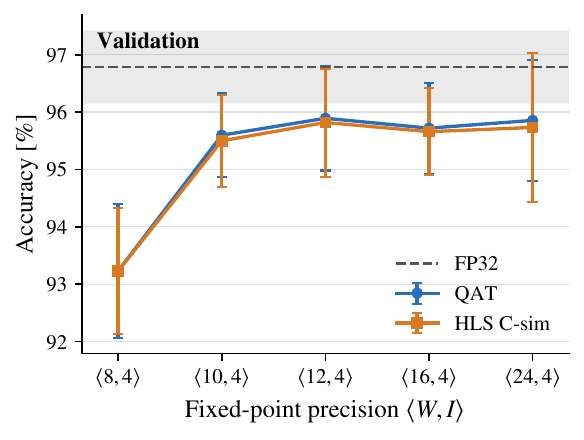}
\end{minipage}\hfill
\begin{minipage}{0.48\linewidth}
\centering
\includegraphics[width=\linewidth]{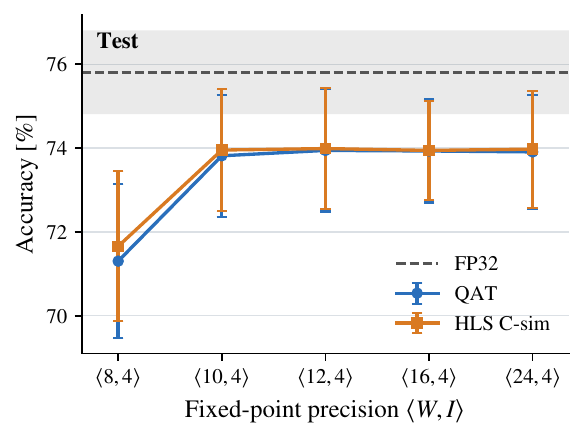}
\end{minipage}
\caption{Validation and test accuracy for the membrane-readout SNN.  The FP32 \pytorch{} reference is shown together with the QAT reference and the generated \hlsfml{} C-simulation result for each included fixed-point precision.
Error bars and shaded regions show uncertainty calculated from $10$-fold cross-validation.}
\label{fig:accuracy-sweep}
\end{figure}

The accuracy results of the precision sweep are shown in Figure~\ref{fig:accuracy-sweep}.
We see that for bit widths of $10$ or larger, we saturate the performance of the quantised SNN on both the validation and the test data.
While the performance of the quantised SNN never reaches that of the FP32 models, it gets close with accuracies of $\sim 96$\%. 
We also see that in all cases the HLS performance is very close to that of the quantised SNNs, indicating a high-fidelity implementation from \pytorch{} to HLS.
In Table~\ref{tab:precision-fidelity} we show the full numerical results for accuracies and for the agreement between the HLS models and the quantised SNNs defined in \pytorch{}. 
The agreement for all models is very high, reaching $\gtrsim 98$\% for bit widths larger than $10$.

\begin{table}[h]
\centering
\footnotesize
\begin{tabular}{lccccc}
\toprule
Precision & QAT val. & HLS val. & QAT test & HLS test & Agreement \\
\midrule
\texttt{<8,4>}  & $93.23\pm1.17\%$ & $93.23\pm1.10\%$ & $71.30\pm1.84\%$ & $71.66\pm1.79\%$ & $97.69\pm0.84\%$ \\
\texttt{<10,4>} & $95.60\pm0.73\%$ & $95.50\pm0.81\%$ & $73.82\pm1.46\%$ & $73.95\pm1.46\%$ & $98.88\pm0.27\%$ \\
\texttt{<12,4>} & $95.89\pm0.91\%$ & $95.82\pm0.94\%$ & $73.94\pm1.46\%$ & $73.99\pm1.44\%$ & $99.65\pm0.17\%$ \\
\texttt{<16,4>} & $95.72\pm0.79\%$ & $95.66\pm0.76\%$ & $73.93\pm1.23\%$ & $73.94\pm1.19\%$ & $99.74\pm0.21\%$ \\
\texttt{<24,4>} & $95.86\pm1.05\%$ & $95.73\pm1.30\%$ & $73.91\pm1.35\%$ & $73.98\pm1.39\%$ & $99.66\pm0.30\%$ \\
\bottomrule
\end{tabular}
\caption{Ten-fold precision sweep for the membrane-readout SNN.  QAT is the fake-quantised \pytorch{} reference.  HLS is \hlsfml{} C simulation.  Agreement is measured between HLS C simulation and the fake-quantised reference on the SHD test set.}
\label{tab:precision-fidelity}
\end{table}

\subsection{Device utilisation and latency}

\begin{figure}[htbp]
\centering
\includegraphics[width=0.74\linewidth]{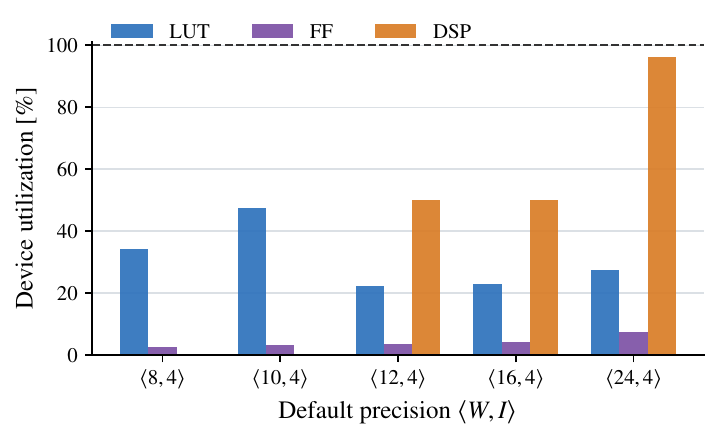}
\caption{Device utilisation after Vivado synthesis for the bit width sweep.}
\label{fig:device-utilisation}
\end{figure}

Now we turn to the device utilisation and latency of SNN inference on FPGAs after synthesis via \hlsfml{} and Vitis/Vivado.
Figure~\ref{fig:device-utilisation} shows Vivado-synthesis device utilisation for the HLS config parameters outlined in the previous section.
We see that hardware utilisation clearly increases as we increase the bit width, and while the lowest precision SNNs use no DSPs, at the high precision end we see that the DSP usage in the device is almost saturated.
As the DSP usage turns on, we also see a decrease in the LUTs used by the SNNs on the hardware.
The FFs remain low throughout, increasing slightly as the precision is increased.
This is a promising result, as we've seen that the validation and test accuracy performance saturate at bit widths of $10$, therefore quantising the network results in big savings on device utilisation while keeping accuracy high. 

In Table~\ref{tab:hls-hardware-summary} we have a more detailed breakdown of the hardware statistics.  
We also include a Latency strategy run with a \texttt{RF} of $1$ for comparison.
The resource numbers are from Vivado synthesis, while the clock period, latency cycles, and initiation interval are from HLS synthesis.  
The full-window latency is computed as the per-timestep maximum HLS latency multiplied by the 140-timestep input window and the requested target clock period.  
For the runs with \texttt{RF} at $8$, the top function has a maximum latency of 15 cycles per timestep, so the reported full-window latency is $15\times140\times16~\mathrm{ns}=33.6~\mu\mathrm{s}$.
We also include a low-precision (\texttt{ap\_fixed<8,4>}) Latency strategy run with an \texttt{RF} of $1$ here for comparison.
While the Latency run achieves a lower clock period, the optimisation in \hlsfml{} assigns much higher LUT resources than are assigned in the Resource strategy runs, and assigns no BRAM or DSPs.
This results in a much larger Initiation Interval and max cycles, leading then to a larger full-window latency.

\begin{table}[htbp]
\centering
\setlength{\tabcolsep}{2.0pt}
\begin{tabular}{lcccccccccccc}
\toprule
Prec. & Strat. & RF & Target & Est. & LUT & FF & BRAM & DSP & Cyc. & II & Win. \\
 & & & ns & ns & & & & & max & & $\mu$s \\
\midrule
\texttt{<8,4>}  & Resource & 8 & 16.0 & 13.03  & 78530  & 11722 & 5.5 & 0    & 15 & 12 & 33.60 \\
\texttt{<10,4>} & Resource & 8 & 16.0 & 13.03  & 109548 & 14276 & 6.0 & 0    & 15 & 12 & 33.60 \\
\texttt{<12,4>} & Resource & 8 & 16.0 & 13.03  & 51653  & 15744 & 5.5 & 864  & 15 & 12 & 33.60 \\
\texttt{<16,4>} & Resource & 8 & 16.0 & 13.109 & 53056  & 19070 & 6.0 & 864  & 15 & 12 & 33.60 \\
\texttt{<24,4>} & Resource & 8 & 16.0 & 13.109 & 63132  & 34037 & 6.0 & 1662  & 15 & 12 & 33.60 \\
\texttt{<8,4>}  & Latency  & 1 & 7.0  & 5.737  & 160299 & 44600 & 0.0 & 0   & 54 & 24 & 52.92 \\
\bottomrule
\end{tabular}
\caption{Hardware results for the membrane-readout SNN on \texttt{xczu7ev-ffvc1156-2-e}.  Resource counts are from Vivado synthesis while clock, latency cycles, and initiation interval are from HLS synthesis.  Full-window latency uses the requested target clock period. }
\label{tab:hls-hardware-summary}
\end{table}

\subsection{SNN workload metrics}

Although the HLS implementation is clock-driven, it is still interesting to study the sparsity of the neuron firing through the network.  
Table~\ref{tab:neurobench-workload} reports workload metrics computed using the \neurobench{} package on the SHD test set for the FP32 model and the included QAT precisions.  
The means and standard deviations here are calculated using the same $10$-fold cross-validation from the previous sections.
The input sparsity is a feature of the SHD dataset, it says that after pooling from $700$ dimensions to $70$, $84$\% of the inputs are zeros.
The hidden sparsity tells us the sparsity in the firing rate of the spiking neurons from the hidden LIF layer, a $0$ would imply that all neurons fire on every update, while a $1$ would imply that none of the neurons ever fire.
Therefore these numbers imply that the hidden layer firing is indeed very sparse, with neurons remaining silent $\sim92$\% of the time.
The connection sparsity is the percentage of weights in the SNN that are numerically $0$.
It makes sense that this number is largest for the low-precision SNNs and gets smaller as the precision increases.

Including all weights, biases, and spiking neuron parameters, \neurobench{} computes that the SNN model here has $29.54\times10^3$ parameters, and that without exploiting any sparsity in the data  the number of dense synaptic-operations required per event inference is $806.4\times10^3$.
The footprint and dense synaptic-operations are static workload metrics, whereas the effective accumulate and effective MAC operations take into account the sparsity in the network with respect to a test sample.
From the table we can see that when sparsity is accounted for, the effective accumulate count is only $(1.1$--$1.4)\times10^4$ operations per sample, and the effective MACs are $\sim1\times10^5$, compared with the full $8.06\times10^5$ dense synaptic operations.  
This means that on dedicated neuromorphic hardware that can capitalise on this sparsity, there would be close an order of magnitude of savings in the power consumed by the SNN vs an ANN.
On the FPGA however, we are not making any savings due to sparsity in the network.
It may be possible to benefit from some of these savings in the future through implementing gating operations in \hlsfml{} to prevent MACs where a spiking neuron does not fire.

\begin{table}[htbp]
\centering
\setlength{\tabcolsep}{3.0pt}
\begin{tabular}{lccccc}
\toprule
Precision & Input sparse & Hidden sparse & Conn. sparse & Eff. ACs & Eff. MACs \\
 & \% & \% & \% & $10^3$ & $10^3$ \\
\midrule
FP32             & $83.98$ & $92.06\pm0.81$ & $0.00\pm0.00$ & $14.24\pm1.45$ & $100.49\pm0.00$ \\
\texttt{<8,4>}  & $83.98$ & $92.47\pm1.12$ & $8.65\pm0.81$ & $10.96\pm1.64$ & $91.65\pm0.89$ \\
\texttt{<10,4>} & $83.98$ & $92.76\pm0.62$ & $2.24\pm0.19$ & $12.26\pm1.03$ & $98.21\pm0.29$ \\
\texttt{<12,4>} & $83.98$ & $92.55\pm0.89$ & $0.62\pm0.10$ & $13.14\pm1.55$ & $99.87\pm0.09$ \\
\texttt{<16,4>} & $83.98$ & $92.51\pm0.81$ & $0.04\pm0.04$ & $13.42\pm1.44$ & $100.45\pm0.04$ \\
\texttt{<24,4>} & $83.98$ & $92.64\pm0.88$ & $0.00\pm0.00$ & $13.19\pm1.57$ & $100.49\pm0.00$ \\
\bottomrule
\end{tabular}
\caption{\neurobench{} workload metrics for the membrane-readout SNN on the SHD test set.  Reported metrics are means and standard deviations over 10 folds.  \neurobench{} also reports accuracy metrics, but these matched the QAT results from the previous sections exactly.}
\label{tab:neurobench-workload}
\end{table}

\section{Discussion \& conclusions}
\label{sec:conclusions}

We have presented and tested an extension of the \hlsfml{} framework enabling SNN models to be compiled to HLS and synthesised and optimised for FPGA deployment through the Vitis / Vivado backend.
This work allows SNN models trained via a \pytorch{} / \snntorch{} workflow to be deployed in existing FPGA-based scientific instruments.  
It provides a route to compare FP32, QAT, and generated-HLS SNN behaviour.  
The current implementation focuses on the LIF neuron, but this work opens a path for future SNN additions to \hlsfml{}, including spike-sparse kernels, recurrent SNNs, convolutional SNNs, and variable-length sequence boundaries.
The SNN inference is clock-driven, so it is scheduled by the FPGA clock and HLS dataflow rather than by asynchronous spike events.  
This means that some advantages typically associated with event-driven neuromorphic systems, such as dynamic avoidance of computation when no spikes are present, are not automatically realized.  
But in future work it may be possible to introduce gate operations that prevent unnecessary computations being performed when a spiking neuron does not spike.
This will not improve latency on the FPGA, but could reduce power consumption for sparsely firing neurons.

The workflow is tested using the SHD benchmark dataset, widely used in the SNN literature, and the results validate the SNN support from HLS right through to the hardware simulation.
The tests spanned both membrane and spike-count outputs from the SNN, and covered a range of bit width precisions.
We show that each SHD event with a $140$ timstep window, with $70$ features per timestep, can be streamed and processed by the SNN in $33.6\mu$s, with large reductions in FPGA resource utilisation through quantising the network to \texttt{ap\_fixed<10,4>} precision while maintaining high accuracy.

Future work will include two strands.  The first is to enrich the SNN functionality in \hlsfml{} by introducing support for more spiking neurons, introduce variable-window capability, and to develop and test more complex machine-learning algorithms in the framework.  For example real-time anomaly detection and reinforcement learning with SNNs via \hlsfml{}.
The second strand is in the application of SNNs to scientific problems that require low-latency real-time machine-learning solutions on time-series data.
This also expands the scope beyond classification to anomaly detection, regression, and reinforcement learning.
To summarise, these results establish a practical baseline for deploying SNNs in scientific FPGA workflows that can be built upon, and provides a foundation for more neuromorphic SNN functionality in future \hlsfml{} releases.

\section*{Acknowledgments}

The author thanks the \hlsfml{} community for the open-source software and documentation, and the \hlsfml{} developers who helped with merging the code.  The author thanks Aqib Javed and Jim Harkin for their collaboration on SNNs and their comments on this project.

\section*{Data \& code}

The SHD data set is publicly available from the Heidelberg Spiking Data Sets project \cite{Cramer_2022}.  
The SNN functionality is deployed in the \hlsfml{} repo at \url{https://github.com/fastmachinelearning/hls4ml}.
The code specific to this SHD analysis is available at \url{https://github.com/bmdillon/snn-shd-hls4ml}.

\bibliographystyle{unsrt}
\bibliography{refs_v3}

\end{document}